\documentclass[preprints,article,accept,pdftex,moreauthors]{Definitions/mdpi} 

\firstpage{1} 
\pubvolume{1}
\issuenum{1}
\articlenumber{0}
\pubyear{2022}
\copyrightyear{2022}
\datereceived{} 
\dateaccepted{} 
\datepublished{} 
\hreflink{https://doi.org/} 

\usepackage{placeins} 
\usepackage{rotating} 

\usepackage{siunitx}
\makeatletter
\newcommand\footnoteref[1]{\protected@xdef\@thefnmark{\ref{#1}}\@footnotemark}
\makeatother

\Title{TorchEsegeta: Framework for Interpretability and Explainability of Image-based Deep Learning Models}

\TitleCitation{TorchEsegeta: Framework for Interpretability and Explainability of Image-based Deep Learning Models}

\Author{Soumick Chatterjee~$^{1,2,3}$*\orcidA{}, Arnab Das~$^{1}$, Chirag Mandal~$^{1}$, Budhaditya Mukhopadhyay~$^{1}$, Manish Vipinraj~$^{1}$, Aniruddh Shukla~$^{1}$, Rajatha Nagaraja Rao~$^{1}$, Chompunuch Sarasaen~$^{3,4}$, Oliver Speck~$^{3,5,6,7}$, and Andreas N{\"u}rnberger~$^{1,2,6}$}

\AuthorNames{Soumick Chatterjee, Arnab Das, Chirag Mandal, Budhaditya Mukhopadhyay, Manish Vipinraj, Aniruddh Shukla, Rajatha Nagaraja Rao, Chompunuch Sarasaen, Oliver Speck, and Andreas N{\"u}rnberger}

\AuthorCitation{Chatterjee, S.; Das, A.; Mandal, C.; Mukhopadhyay, B.; Vipinraj, M.; Shukla, A.; Rao, R.N.; Sarasaen, C.; Speck, O.; N{\"u}rnberger, A.}

\address{%
$^{1}$ \quad Faculty of Computer Science, Otto von Guericke University Magdeburg, Germany\\
$^{2}$ \quad Data and Knowledge Engineering Group, Otto von Guericke University Magdeburg, Germany\\
$^{3}$ \quad Biomedical Magnetic Resonance, Otto von Guericke University Magdeburg, Germany\\
$^{4}$ \quad Institute for Medical Engineering, Otto von Guericke University Magdeburg, Germany\\
$^{5}$ \quad German Center for Neurodegenerative Disease, Magdeburg, Germany\\
$^{6}$ \quad Center for Behavioral Brain Sciences, Magdeburg, Germany\\
$^{7}$ \quad Leibniz Institute for Neurobiology, Magdeburg, Germany}

\corres{Correspondence: soumick.chatterjee@ovgu.de;}

\abstract{Clinicians are often very sceptical about applying automatic image processing approaches, especially deep learning based methods, in practice. One main reason for this is the black-box nature of these approaches and the inherent problem of missing insights of the automatically derived decisions. In order to increase trust in these methods, this paper presents approaches that help to interpret and explain the results of deep learning algorithms by depicting the anatomical areas which influence the decision of the algorithm most. Moreover, this research presents a unified framework, TorchEsegeta, for applying various interpretability and explainability techniques for deep learning models and generate visual interpretations and explanations for clinicians to corroborate their clinical findings. In addition, this will aid in gaining confidence in such methods. The framework builds on existing interpretability and explainability techniques that are currently focusing on classification models, extending them to segmentation tasks. In addition, these methods have been adapted to 3D models for volumetric analysis. The proposed framework provides methods to quantitatively compare visual explanations using infidelity and sensitivity metrics. This framework can be used by data scientists to perform post-hoc interpretations and explanations of their models, develop more explainable tools and present the findings to clinicians to increase their faith in such models. The proposed framework was evaluated based on a use case scenario of vessel segmentation models trained on Time-of-fight (TOF) Magnetic Resonance Angiogram (MRA) images of the human brain. Quantitative and qualitative results of a comparative study of different models and interpretability methods are presented. Furthermore, this paper provides an extensive overview of several existing interpretability and explainability methods.}


\begin{document}

\section{Introduction}
\label{ch:introduction}

The use of artificial intelligence is widely prevalent today in medical image analysis. It is, however, imperative to do away with the black-box nature of Deep Learning techniques to gain the trust of radiologists and clinicians, as a model's erroneous output might have a high impact in the medical domain. 

\textit{Interpretability vs Explainability: }
Interpretability and Explainability methods aim to unravel the black-box nature of decision-making in machine learning and deep learning models. Explainability focuses on explaining the internal working mechanisms of the model. In contrast, interpretability focuses on observing the effects of changes in model parameters and inputs on the model prediction, hence attributing properties to the input, and it also requires a context, considering the audience who is to interpret the properties and characteristics for the model outcomes~\citep{marcinkevivcs2020interpretability,chakraborty2017interpretability}. It is to be noted that earlier work failed to draw a clear line of distinction between interpretability and explainability as these terms are subjective and pertain to the stakeholders, considering the audience who must understand the model and its outcomes. Interpretability and explainability nurture a sense of human-machine trust as they help the users of the machine/deep-learning models understand how certain decisions are made by the model and are not limited to statistical metric-based approaches such as accuracy or precision. This in turn, allows employing such models in mission-critical problems such as medicine, autonomous driving, legal systems, banking etc. The basis of measuring model transparency is said to comprise simulatability, decomposability, and algorithmic transparency~\citep{chakraborty2017interpretability}; and model functionality which consists of textual descriptions of the model's output and visualisations of the model's parameters. The goals to be attained through interpretability of models~\citep{marcinkevivcs2020interpretability} are trust, reliability, robustness, fairness, privacy, and causality. Explainability can be formulated as the explanation of the decisions made internally by the model that in turn generates the observable, external conclusions arrived at by the model. This promotes human understanding of the model's internal mechanisms and rationale, which in turn helps build user trust in the model~\citep{emmert2020explainable}. The evaluation criteria for model explainability include comprehensibility by humans, fidelity and scalability of the extracted representations, and scalability and generality of the method~\citep{belle2020principles}. 

Some models are transparent and hence inherently understandable, such as Decision Trees, K-Nearest Neighbours, Rule-based and Bayesian models. There are even some deep learning based models, like GP-UNet~\citep{dubost2017gp} and CA-Net~\citep{gu2020net}, which are inherently trying to provide an explanation concerning their outcome. but some are opaque and require post-hoc explanations. A post-hoc model explanation can be through model-agnostic and model-specific techniques that aim towards explaining a black-box model in human-understandable terms. The European Union has also incorporated transparency and accountability of models in 2016 as a criterion in the ethical guidelines of trustworthy AI~\citep{arrieta2020explainable,choo2018visual}. Through interpretability and explainability, to some degree, the following goals can be achieved: trustworthiness, causality, confidence, fairness, informativeness, transferability and interactivity, which in turn can help improve the model~\citep{arrieta2020explainable}.

Data scientists can generate interpretability-explainability results for their opaque models and present the model's reasoning or the model's inner mechanism to the domain experts (e.g. clinicians). If the reasoning is the same as a human domain expert would have done, then the experts can have more faith in that model, which in turn implies that the model can be incorporated into real-life workflows. Apart from building trust in the models, interpretability and explainability techniques can be used by the data scientists to improve their models as well, by discussing the outcomes with the domain experts and improving the model based on expert feedback. Moreover, accurate (verified by experts) interpretability-explainability results can be used as part of automatic or semi-automatic teaching programs for trainees.  


For classification models, there are multiple interpretability ad explainability techniques supported by various libraries such as Captum\footnote{\url{https:captum.ai}}, Torchray~\citep{fong2019understanding} and CNN Visualization library \footnote{\url{https://github.com/utkuozbulak/pytorch-cnn-visualizations}} and many of the techniques are introduced in the following section of this paper. However, this is more challenging for segmentation as the output is more complex than just a simple class prediction. In this contribution, various interpretability and explainability techniques used for classification models have been adapted to work with segmentation models. On applying the interpretability techniques, essential features or areas of the input image can be visualised, on which the model's output is critically based. By applying explainability techniques, a better understanding of the knowledge represented in the model's parameters can be achieved.

\subsection{Contributions}\label{sec:Contrib}
This paper proposes a unified, flexible and scalable interpretability-explainability pipeline for PyTorch, TorchEsegeta, which leverages post-hoc interpretability and explainability methods and can be applied on 2D or 3D deep learning models working with images. Apart from implementing the existing methods for classification models, this research extends them for segmentation models. This pipeline can be applied to trained models with little to no modification, using either a graphical user interface for easy access or directly using a python script. It also provides an easy platform for incorporating new interpretability or explainability techniques. Moreover, this pipeline provides features to evaluate the interpretability and explainability methods using two different methods. First, using cascading randomisation of the model's weights, which can help evaluate how much the results are dependent on the actual weights, and second, using quantitative metrics: infidelity and sensitivity. Finally, the pipeline was applied on models trained to segment vessels from magnetic resonance angiograms (MRA), and the interpretability results are shown here. Furthermore, apart from the technical contributions with the TorchEsegeta pipeline, this paper also provides a comprehensive overview of several post-hoc interpretability and explainability techniques.



\section{Methods}
\label{ch:methods}
This paper presents the TorchEsegeta framework, which integrates various interpretability and explainability techniques available in different libraries and extends these techniques for segmentation models. It is noteworthy that the development of this pipeline started with the exploration of various interpretability techniques for classifying COVID-19 and other types of pneumoniæ~\citep{chatterjee2020exploration}. An initial pipeline was developed under that project for classification models, but only for 2D images. This research further extends that for 3D volumetric images, as well as for segmentation models. Apart from incorporating these features, the original pipeline was further streamlined and improved during this research - to create the first version of TorchEsegeta.

\subsection{Incorporated libraries} 
The following interpretability and explainability based libraries have been explored in this research work - Captum, CNN visualisation, TorchRay, DeepDream, Lucent, LIME and SHAP 

Captum is a library built on PyTorch and is used to provide the interpretability of machine/deep-learning models. It provides many algorithms that evaluate the contribution of different features in providing a model's prediction and thus helps in improving the model.

CNN visualisations~\citep{uozbulak_pytorch_vis_2019} provides different implementations for the different interpretability techniques and visualisations for CNN based models architectures.

TorchRay \footnote{\url{https://github.com/facebookresearch/TorchRay}} is a package used for several visualisation methods for 
convolutional neural network architectures using PyTorch. It focuses on interpretability wherein it attempts to determine which regions of the input image influence the final prediction made by the model~\citep{fong2019understanding}.

LIME (Local Interpretable Model-agnostic Explanations) is a model-agnostic technique that provides post-hoc model explanations to explain the decisions of the deep learning model, and due to its model-agnostic nature, LIME flexibly explains any unknown model~\citep{choo2018visual, samek2020toward, arrieta2020explainable}

SHAP \footnote{\url{https://christophm.github.io/interpretable-ml-book/shap.html}} (SHapley Additive exPlanation) is a post-hoc, game-theoretical approach that computes shapley values in order to explain the prediction made by the deep learning model~\citep{lundberg2017unified}.

Lucent \footnote{\url{https://github.com/greentfrapp/lucent}} is the PyTorch implementation of lucid for the explainability of deep learning models. It aims to explain the decision made by the deep neural network by explaining what is being learnt by the various layers of the network.

DeepDream \footnote{\url{https://ai.googleblog.com/2015/06/inceptionism-going-deeper-into-neural.html}} is also an explainability technique that is used to visualise the parameters learnt by the convolutional neural network.

\subsection{Implemented Interpretability Techniques}
Interpretability techniques help to understand the focus area of a model - can help understand the reasoning done by the model. These techniques can be categorised into two groups: model attribution and layer attribution.  

\subsubsection{Model Attribution Techniques}

Model attribution techniques are the techniques to assess the contribution of each attribute to the prediction of the model. There is a long list of methods available in the literature under the rubric model attribution techniques, hence for better understanding, they are further divided here into two groups: feature-based and gradient-based.

\begin{enumerate}
\item \textit{Feature-based Techniques:}

Methods under this group bring into play input and/or output feature space to compute local or global model attributions.
\begin{enumerate}

 \item \textit{Feature Permutation} - This is a perturbation based technique ~\citep{breiman2001random,fisher2019all} in which the value of an input or group of inputs is changed utilising random permutation of values in a batch of inputs and calculating its corresponding change in output. Hence, meaningful feature attributions are calculated only when a batch of inputs is provided.

  \item \textit{Shapley Value Sampling} - The proposed method defines an input baseline to which all possible permutations of the input values are added one at a time, and the corresponding output values and hence each feature attribution is calculated. For permutation O, given player set N, the set of all possible permutations $ \pi(N)$ and the predecessors of player i $ Pre^i(O)$, the Shapley value is given by 
  \begin{equation}
      Sh_i (v) = \sigma_{O\in \pi(N)} \frac{1}{n!}(v(Pre^i(O) \bigcup i) - v(Pre^i(O)))
  \end{equation}
  As all possible permutations are considered, this technique is computationally expensive, and this can be overcome by sampling the permutations and averaging their marginal contributions instead of considering all possible permutations such as the ApproShapley sampling technique~\citep{castro2009polynomial}

  \item \textit{Feature Ablation} - This method works by replacing an input, or a group of inputs with another value defined by a range or reference value and the feature attribution of the input or group of inputs such as a segment are computed. This method works based on perturbation.
  
  \item \textit{Occlusion} - Similar to the Feature Ablation method, Occlusion~\citep{zeiler2014visualizing} works as a perturbation based approach wherein the continuous inputs in a rectangular area are replaced by a value defined by a range or a reference value. Using this approach, the change in the corresponding outputs is calculated in order to find the attribution of the feature.

\item \textit {RISE Randomised Input Sampling for
Explanation of Black-box Models} - It generates an importance map indicating how salient each pixel is for the model's prediction~\citep{petsiuk2018rise}. RISE works on black-box models since it does not consider gradients while making the computations. In this approach, the model's outputs are tested by masking the inputs randomly and calculating the importance of the features.

\item \textit {Extremal Perturbations} - They are regions of an image that maximally affect the activation of a certain neuron in a neural network for a given area in an image~\citep{fong2019understanding}. Extremal Perturbations lead to the largest change in the prediction of the deep neural network, when compared to other perturbations defined by 
\begin{equation}
    m_a = \underset{m:||m||_1 a|\Omega|, m\in M}{\arg\max} \Phi(m\otimes x)
\end{equation}
for the chosen area a. The paper also introduces area loss to enforce the restrictions while choosing the perturbations.

  \item \textit {Score-weighted Class Activation (Score CAM)} - It is a gradient-independent interpretability technique based on class activation mapping ~\citep{wang2020score}. Activation maps are first extracted, and each activation then works as a mask on the original image, and its forward-passing score on the target class is obtained. Finally, the result can be generated by the linear combination of score-based weights and activation maps\footnote{\url{https://github.com/haofanwang/Score-CAM}}. Given a convolutional layer l, class c, number of channels k and activations A:
  \begin{equation}
      L_{Score-CAM} ^c = ReLU(\sum_k \alpha_k^c A_l^k)
  \end{equation}
    
\end{enumerate}

\paragraph{Gradient-based Techniques:}
This group of methods mainly use the model's parameter space to generate the attribute maps - calculated with the help of the gradients. 

\begin{enumerate}

 \item \textit{Saliency} - It was initially designed for visualising the image classification done by convolutional networks and the saliency map for a specific image~\citep{simonyan2013deep}. It is used to check the influence of each pixel of the input image in assigning a final class score to the image $I$ using the linear score model $ S_c(I) = w_c^T I + b_c $ for weight w and bias b.
  
  \item \textit{Guided Backpropagation} -
  In this approach, during the backpropagation, the gradients of the outputs are computed with respect to the inputs~\citep{springenberg2014striving}. RELU activation function is applied to the input gradients, and direct backpropagation is performed, ensuring that the backpropagation of non-negative gradients does not occur.

  \item \textit{Deconvolution} - This approach is similar to the guided backpropagation approach wherein it applies RELU to the output gradients instead of the input gradients and performs direct backpropagation~\citep{zeiler2014visualizing}. Similarly, RELU backpropagation is overridden to allow only non-negative gradients to be backpropagated.

  \item \textit{Input X Gradient} - It extends the saliency approach such that the contribution of each input to the final prediction is checked by multiplying the gradients of outputs and their corresponding inputs in a setting of a system of linear equations AX = B where A is the gradients and B is the calculated final contribution of input X ~\citep{shrikumar2016not}.

  \item \textit{Integrated gradients} - This technique ~\citep{sundararajan2017axiomatic} calculates the path integral of the gradients along the straight line path from the baseline ${x^\prime}$ to the input x. It satisfies the axioms of Completeness i.e. the attributions must account for the difference in output for the baseline ${x^`}$ and input x, Sensitivity i.e. a non-zero attribution must be provided even to inputs that flatten the prediction function where the input differs from the baseline, and Implementation Invariance of gradients i.e. two functionally equivalent networks must have identical feature attributions. It requires no instrumentation of the deep neural network for its application. All the gradients along the straight line path from $x^\prime$ to x are integrated along the $i^{th}$ dimension as:
  \begin{equation}
      IG_i(x) ::= (x_i - x_i^{\prime}) \times \int_{\alpha = 0}^{1} \frac{\partial F(x^{\prime} + \alpha \times (x - x^{\prime}))}{\partial x_i} d\alpha
  \end{equation}
  
  \item \textit {Grad Times Image} - In this technique~\citep{shrikumar2017just} the gradients are multiplied with the image itself. It is a predecessor of the DeepLift method as the activation of each neuron for a given input is compared to its reference activation, and contribution scores are assigned according to the difference for each neuron. 

\item \textit{DeepLift} - It is a method~\citep{shrikumar2017learning}  that considers not only the positive but also the negative contribution scores of each neuron based on its activation concerning its reference activation. The difference in the output of the activation function of the neuron t under observation is calculated based on the difference in input with respect to a reference input. Contribution scores for each of the preceding neurons $x_1,..x_n$ that influence t are assigned $C_{{\triangle x_i}_{\triangle t}}$ that sums up to the difference in t's activation:
\begin{equation}
    \sum_{i=1}^{n} C_{{\triangle x_i}_{\triangle t}} = \triangle t
\end{equation}
  
\item \textit{DeepLiftShap} - This method
extends the DeepLift method and computes the SHAP values on an equivalent, linear approximation of the model~\citep{lundberg2017unified}. It assumes the independence of input features. For model f, the effective linearization from SHAP for each component is computed as: 
\begin{equation}
    \phi_i (f_3,y) \approx m_{y_i f_3}(y_i - E[y_i])
\end{equation}
   
  \item \textit{GradientShap} - This technique also assumes the independence of the inputs and computes the game-theoretic SHAP values of the gradients on the linear approximation of the model~\citep{lundberg2017unified}. Gaussian noise is added to randomly selected points, and the gradients of their corresponding outputs are computed.
  
  \item \textit{Guided GradCAM} - The Gradient-weighted Class Activation Mapping approach provides a means to visualise the regions of an image input that are predominant in influencing the predictions made by the model. This is done by visualising the gradient information pertaining to a specific class in any layer of the user's choice. It can be applied to any CNN-based model architecture. The guided backpropagation and GradCAM approaches are combined by computing the element-wise product of guided backpropagation attributions with upsampled GradCAM attributions~\citep{selvaraju2017grad}.

  \item \textit {Grad-Cam++} - Generalised Gradient-based Visual Explanations for Deep Convolutional Networks is a method~\citep{chattopadhay2018grad} that claims to provide better predictions than the Grad-CAM and other state-of-the-art approaches for object localisation and explaining occurrences of multiple object instances in a single image. This technique uses a weighted combination of the positive partial derivatives of the last convolutional layer's feature maps concerning a specific class score as weights to generate a visual explanation for the corresponding class label. The importance $w_k^c$ of an activation map $ A^k$ over class score $ Y^c$is given by 
  \begin{equation}
      w_k^c = \sum_i \sum_j \alpha_{ij}^{kc}.relu(\frac{\partial Y^c}{\partial A_{ij}^k})
  \end{equation}
  Grad-CAM++ provides human-interpretable visual explanations for a given CNN architecture across multiple tasks, including classification, image caption generation and 3D action recognition.

\item \textit {Vanilla Backpropagation and layer visualisation} - It is the standard gradient backpropagation technique through the deep neural network wherein the gradients are visualised at different layers and what is being learnt by the model is observed, given a random input image.

 \item \textit {Smooth Grad} - In this method~\citep{wang2020score,smilkov2017smoothgrad} random Gaussian noise $ N(0,\sigma^2)$ is added to the given input image and the corresponding gradients are computed to find the average values over n image samples 
\begin{equation}
    \hat{M}_{c}(x) = \frac{1}{n} \sum_{1}^{n}M_c(x + N(0,\sigma^2))
\end{equation}
 
 Vanilla and guided backpropagation techniques can be used to calculate the gradients.

\end{enumerate}
\end{enumerate}

\subsubsection{Layer Attribution Techniques}
Layer Attribution methods evaluate the effect of each individual neuron in a particular layer on the model output. There are various types of layer attribution methods that have been explored as part of this research work. Perturbation-Based approaches~\citep{wang2020score}  perturb the original input and observe the change in the prediction of the model, and are highly time-consuming.
 
 \begin{enumerate}

\item \textit {Inverted representation} - The aim of this technique is to generate the given input image after a number of n target layers. An inverse of the given input image representation is computed~\citep{mahendran2014understanding} in order to find an image $\Phi (x_0)$ whose representation best matches the input image $\Phi _0$ while minimising the given loss l such that 
\begin{equation}
    x^* = {\underset {x \in { \mathbb {R}}^{HxWxC}}{\arg\min} \hspace{0.1cm} l(\Phi(x), \Phi _0) + \lambda R(x)}
\end{equation}

\item \textit {Layer Activation with Guided Backpropagation} - This method~\citep{springenberg2015striving} is quite similar to guided backpropagation, but instead of guiding the signal from the last layer and a specific target, it guides the signal from a specific layer and filter. The guided backpropagation method adds an additional guidance signal from the higher layers to the usual backpropagation.

\item \textit {Layer DeepLift} - This is the DeepLift method as mentioned in the model attribution techniques~\citep{lundberg2017unified}, but applied with respect to the particular hidden layer in question.

\item \textit {Layer DeepLiftShap} - This method is similar to the DeepLIFT SHAP technique mentioned in the model attribution techniques~\citep{lundberg2017unified}, applied for a particular layer. The original distribution of baselines is taken, and the attribution for each input-baseline pair is calculated using the Layer DeepLIFT method, and the resulting attributions are averages per input example. Assuming model linearity, $\phi(f_3,y) \approx m_yi$

\item \textit {Layer GradCam} – The GradCam attribution for a given layer is provided by this method~\citep{selvaraju2017grad}. The target output's gradients are computed concerning the particularly given layer. The resultant gradients are averaged for each output channel (dimension 2 of output). The average gradient for each channel is then multiplied by the layer activations. The results are then added over all the channels. For the class feature weights w, global average pooling is performed over the feature maps A such that 
\begin{equation}
    S^c = \sum_{k} w_k^c \frac{1}{Z} \sum_{i} \sum_{j} A_{ij}^k
\end{equation}

\item \textit {Layer GradientShap} - This is analogous with GradientSHAP ~\citep{lundberg2017unified} method as mentioned in the model attribution techniques but applied for a particular layer. Layer GradientSHAP adds Gaussian noise to each input sample multiple times, wherein a random point along the path between baseline and input is selected, and the gradient of the output with respect to the identified layer is computed. The final SHAP values approximate the expected value of gradients * (layer activation of inputs - layer activation of baselines).

\item \textit {Layer Conductance} – This method~\citep{dhamdhere2018important} provides the conductance of all neurons of a particular hidden layer. Conductance of a particular hidden unit refers to the flow of Integrated Gradients attribution through this hidden unit. The main idea behind using this method is to decompose the computation of the Integrated Gradients via the chain rule. One property that this method of conductance satisfies is that of completeness. Completeness means that the conductances of a particular layer add up to the prediction difference of F(x) – F($x^{\prime}$) for input x and baseline input $x^{\prime}$. Other properties that are satisfied by this method are that of Linearity and insensitivity.

\item \textit {Internal Influence} - This method calculates the contribution of a layer to the final prediction of the model by integrating the gradients with respect to the particular layer under observation. The internal representation is influenced by an element j as defined by ~\citep{leino2018influence} such that 
\begin{equation}
     \chi _i^s(f,P) = \int_{\chi} \frac{\partial g}{\partial z_j}\Big|_{h(x)} P(x)dx
\end{equation}
wherein $s = \langle g,h \rangle $ for the slice of the network represented by s as a function of tuple g,h. It is similar to the integrated gradients approach where instead of the input, the gradients are integrated with respect to the layer.

\item \textit {Contrastive Excitation Backpropagation/Excitation Backpropagation} - This approach is used to generate and visualise task-specific attention maps. Excitation Backprop method as proposed by~\citep{zhang2018top} is to pass along top-down signals downwards in the network hierarchy via a probabilistic Winner-Take-All process wherein the most relevant neurons in the network are identified for a given top-down signal. Both top-down and bottom-up information is integrated to compute the winning probability of each neuron as defined by  
\begin{equation}
    y_i = \sum_{j=1}^N w_{ij} x_j
\end{equation}
for input x and weight matrix w. The contrastive excitation backpropagation is used to make the top-down attention maps more discriminative.

\item \textit {Layer Activation} – It computes the activation of a particular layer for a particular input image~\citep{liu2020evolving}. It helps to understand how a given layer reacts to an input image. One can get an excellent idea of what part or features of the image at which a particular layer looks. 

\item \textit {Linear Approximator} - This is a technique to overcome inconsistencies of post-hoc model interpretation; linear approximator combines a piecewise linear component and a nonlinear component~\citep{guo2020interpretable, fong2019understanding}. The piecewise linear component describes the explicit feature contributions by piecewise linear approximation that increases
the expressiveness of the deep neural network. The nonlinear component uses a multi-layer perceptron to capture feature interactions and implicit nonlinearity, which in turn increases the prediction performance. Here, the interpretability is obtained once the model is learned in the form of feature shapes and has high accuracy.

\item \textit {Layer Gradient X Activation} – This method~\citep{ancona2017towards} computes the element-wise product of a given layer’s gradient and activation. It is a combination of the gradient and activation methods of layer attribution. The output attributions are returned as a tuple if the layer input/output contains multiple tensors or a single tensor is returned.

\end{enumerate}

\subsection{Implemented Explainability Techniques}
Explainability techniques aim to unravel the internal working mechanism of a model - tries to explain the knowledge represented inside the model's parameters. 

\subsubsection{DeepDream}
As the network is trained using many examples, it is essential to check what has been learnt from the input image. Hence, visualising what has been learnt at different layers of the CNN based network gives rise to repetitive patterns of different levels of abstraction, enabling to interpret of what has been learnt by the layers. This visualisation can be realised using DeepDream. Given an arbitrary input image, any layer from the network can be picked, and the detected features at that layer can be enhanced. It is observed that the initial layers are sensitive to basic features in the input images, such as edges, and the deeper layers identify complex features from the input image. As an example, InceptionNet \footnote{https://ai.googleblog.com/2015/06/inceptionism-going-deeper-into-neural.html} was trained on animal images, and it was observed that the different layers of the network could interpret an interesting remix of the learnt animal features in any given image.

\subsubsection{LIME}

Local Interpretable Model-agnostic Explanations (LIME) is a technique for providing post-hoc model explanations. LIME constructs an approximated surrogate, locally linear model or decision tree of a given complex model that helps to explain the decisions of the original model, and due to its model-agnostic nature, it can be employed to explain any model even when its architecture is unknown~\citep{choo2018visual, samek2020toward, arrieta2020explainable}. LIME also performs the transformation of input features to obtain a representation that is interpretable to humans~\citep{belle2020principles}. In the original work~\citep{ribeiro2016should}, the authors propose LIME, an algorithm that provides explanations for a model f, that are locally faithful in the locality $\Pi_x$, for an individual prediction of the model, such that users can ensure that they can trust the prediction before acting on it. LIME provides an explanation for f in the form of a model g and  $g \in G$ where G is a set of all possible interpretable models and $\Omega(g)$ which is the complexity of the interpretable model, is minimised. $L(f,g, \Pi_x )$ is the fidelity function that measures the unfaithfulness of g in estimating f in the locality $\Pi_x$. Hence, the explanation provided by LIME is~\citep{ribeiro2016should}:

\begin{equation}
\xi(x) = argmin_{g\in G} L(f,g, \Pi_x ) + \Omega(g)
\end{equation}

\subsubsection{SHAP}
SHapley Additive exPlanation~\citep{lundberg2017unified} is similar to LIME such that it approximates an interpretable, explanation model g of the original, complex model f, in order to explain a prediction made by the model f(x). SHAP provides post-hoc model explanations for an individual output and is model-agnostic. It calculates the contribution of each feature in producing the final prediction and is based on the principles of Game Theory such as Shapley Values~\citep{belle2020principles}. SHAP works on simplified, interpretable input data $x^`$, which is analogous to the original input data such that $x = h_x(x)$. SHAP must ensure consistency theorems~\citep{lundberg2017unified},local accuracy such that $g(x^`)$ matches $f(x)$ when $x = h_x(x)$ and missingness such that if an attribute's value is 0, it does not not imply that its corresponding contribution to the prediction is 0. Hence, the contribution of each attribute x can be calculated as follows~\citep{lundberg2017unified}:

\begin{equation}
\phi_i(f,x) = \Sigma _{{z^`} \subseteq {x^`}} \frac{|z^`|! (M - |z^`| - 1)!}{M!} [f_x(z^`) f_x(z^`\setminus i)]
\end{equation}
where $z^` \in \{0,1\}^M$ and $|z^`|$ is the number of non-zero entries in $z^`$ and $z^` \subseteq x^`$ represents all $z^`$  vectors where the non-zero entries are a subset of the non-zero entries in $x^`$.

\subsubsection{Lucent}
Based on Lucid \footnote{\url{https://github.com/tensorflow/lucid}}, Lucent is a library implemented using PyTorch for the explainability of deep learning models. The implementation provides optivis, which is the main framework for providing the visualisations of parameters learnt by the different layers of the deep neural network. It can be used to visualise torchvision models with no overhead for setup. Activation atlas methods, feature visualisation methods, building blocks and differentiable image parameterisations can be used to visualise the different features learnt by the network. Activation atlas methods comprise different methods which show the network activations for a particular class or average activations in a grid cell of the image. Feature visualisation methods help understand the crucial features for a neuron, entire channel, or layer. Building blocks help to visualise the activation vector and its components for the given image. Differentiable image parameterisations find the types of image generation processes which can be backpropagated through, and this helps to perform appropriate preconditioning of the input image to improve the optimisation of the neural network. 

\subsection{The Pipeline: TorchEsegeta} 

\begin{figure}
			\centering
			\includegraphics[width=0.9\textwidth]{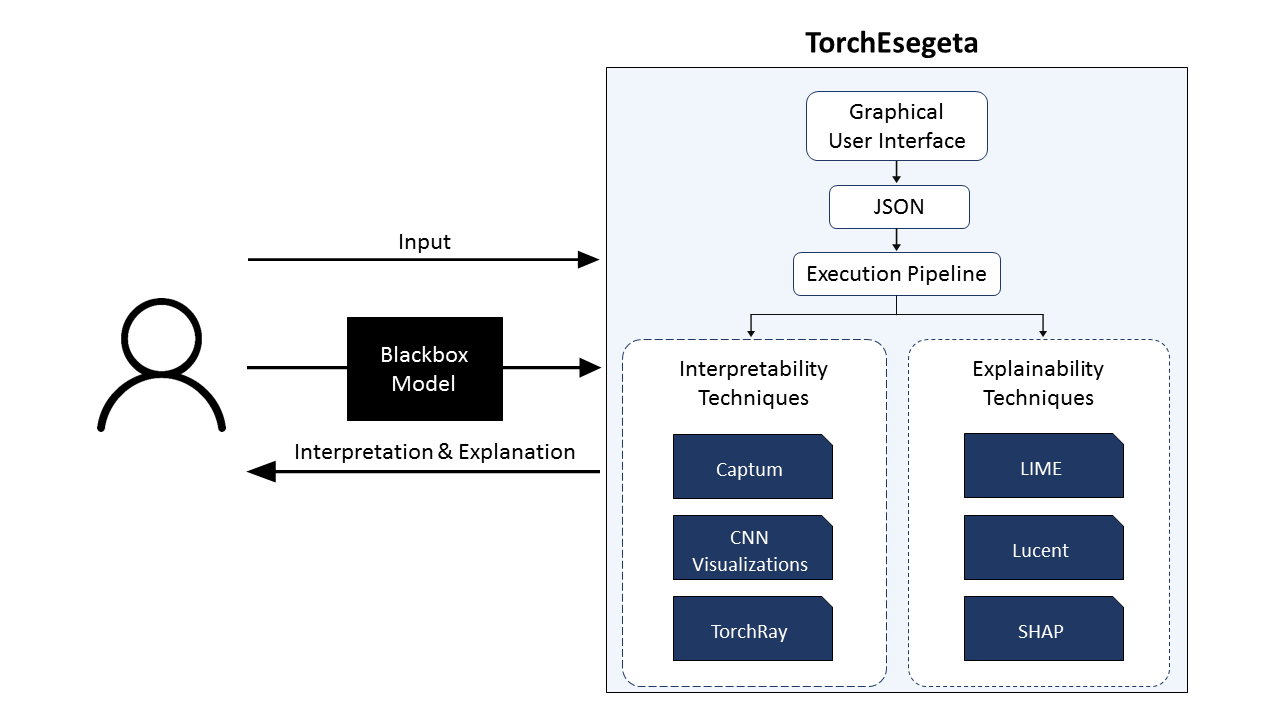}
			\captionsetup{justification=centering}
            \caption{TorchEsegeta pipeline architecture}
	        \label{fig:pipe}
   \end{figure}

The pipeline architecture is shown in Figure \ref{fig:pipe}.
The pipeline is implemented keeping in line with the object-oriented methodology. One of the key features of the pipeline is that it is easily scalable, according to the customised needs of the end-user because of the JSON based configuration. The pipeline is a plug and play software and can be easily used by the end-users. This pipeline is publicly available on GitHub\footnote{\label{git}TorchEsegeta on GitHub: \url{https://github.com/soumickmj/TorchEsegeta}}.

\subsection{Features of TorchEsegeta} 

3D input data is especially important for medical images like MRI and CT scans. Hence, the framework has been built to support both 2D and 3D input data. One of the most important features of the pipeline is scalability. New methods can be added to the pipeline seamlessly by the users and used as per requirement. 

The JSON based configuration will allow the user to execute the pipeline according to their specific needs and is easy to use the feature. The logging facility allows the users to track the pipeline execution and make changes in case of any error because of wrong parameter usage.

The timeout facility makes sure that no method execution is above a certain threshold of time. That way, it saves both computation resources as well as valuable time for the end-user.

\subsubsection{Parallel Execution}

 Another way of saving much valuable time for the user is by using Multi-GPU, which helps in the parallel execution of multiple methods simultaneously. The multi-threading feature enables the execution of multiple methods on the same GPU allowing the users to make optimum use of resources. 

\subsubsection{Patch-based Execution}
Patch-based models are supported in the pipeline and thus helps running computationally intensive methods. Existing interpretability and explainability methods do not give an option to be directly used on patch-based models.

\subsubsection{Automatic Mixed Precision}
Automatic Mixed Precision~\citep{micikevicius2017mixed} facility is also incorporated in the framework aiding in the faster execution of the code. Due to the use of this technique, the overall memory consumption while executing the methods is greatly reduced. The execution time is also reduced as a result.

\subsubsection{Wrapper for Segmentation Models}
The interpretability techniques that are available publicly are basically for classification problems. The framework is an extension to the segmentation problem with the help of a task-specific wrapper functionality - 
\begin{enumerate}
    \item Pixel-wise multi-class classification
    \item Threshold-based pixel classification
\end{enumerate}
\paragraph{Pixel-wise multi-class classification} In this method, the pixel scores are summed up for all pixels predicted as each class. Two main steps are performed:

a. Pixel-wise Class Assignment - In this step, for every pixel, the argmax class is computed.
 Let y be the image with dimensions 
 \begin{equation}
  y_{ij} = \underset{k}{\arg\max}(y_{ijk})   
 \end{equation}

 b. Final Output Calculation - In this step, the sum of the pixel scores for all pixels is predicted as each class is computed. 
  \begin{equation}
  Out_{in} = count\{y_{ij} \hspace{0.2cm}\big | \hspace{0.2cm} y_{ij} \in class \hspace{0.2cm}m \} 
  \end{equation}
  where $0 \leq m \leq N_c$
  
 \paragraph{Threshold based pixel classification} This method performs class identification by Otsu threshold and then sums up the pixels for each class. This task is also performed in two steps:

a. Normalisation - In this step, the input image is normalised by the following function:\\
\begin{equation}
y_{ij_{norm}} = \frac{y_{ij} - min(y)}{max(y) - min(y)}
\end{equation}

b. Pixel-wise binarisation - The pixel-wise binarisation is performed with the help of Otsu thresholding.

\begin{equation}
y_{ij} = \Bigg\{ ^{1\hspace{0.2cm}where\hspace{0.2cm}y_{ij_{norm}} > th } _{0\hspace{0.2cm} elsewhere}
\end{equation}
where th = otsu($y_{ij_{norm}}$)

The output for both the processes is a tensor with a single value for each class.

\subsubsection{Graphical User Interface}
\label{ch:gui}
\begin{figure}
			\centering
			\includegraphics[width=0.6\textwidth]{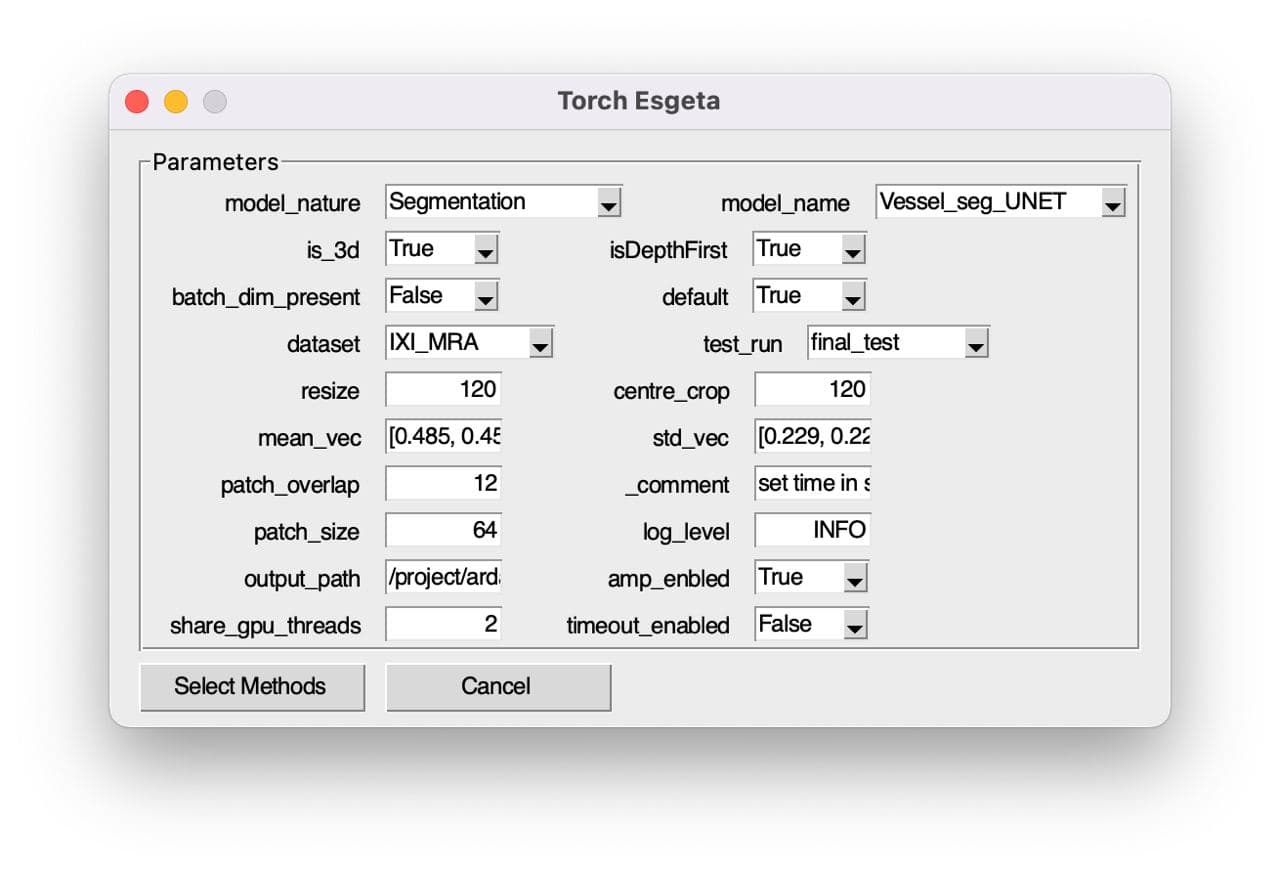}
			\captionsetup{justification=centering}
            \caption{GUI: General parameter selection. Select the desired common parameters for running the pipeline, including the selection of the model.}
	        \label{fig:UI_1}
   \end{figure}

\begin{figure}
			\centering
			\includegraphics[width=\textwidth]{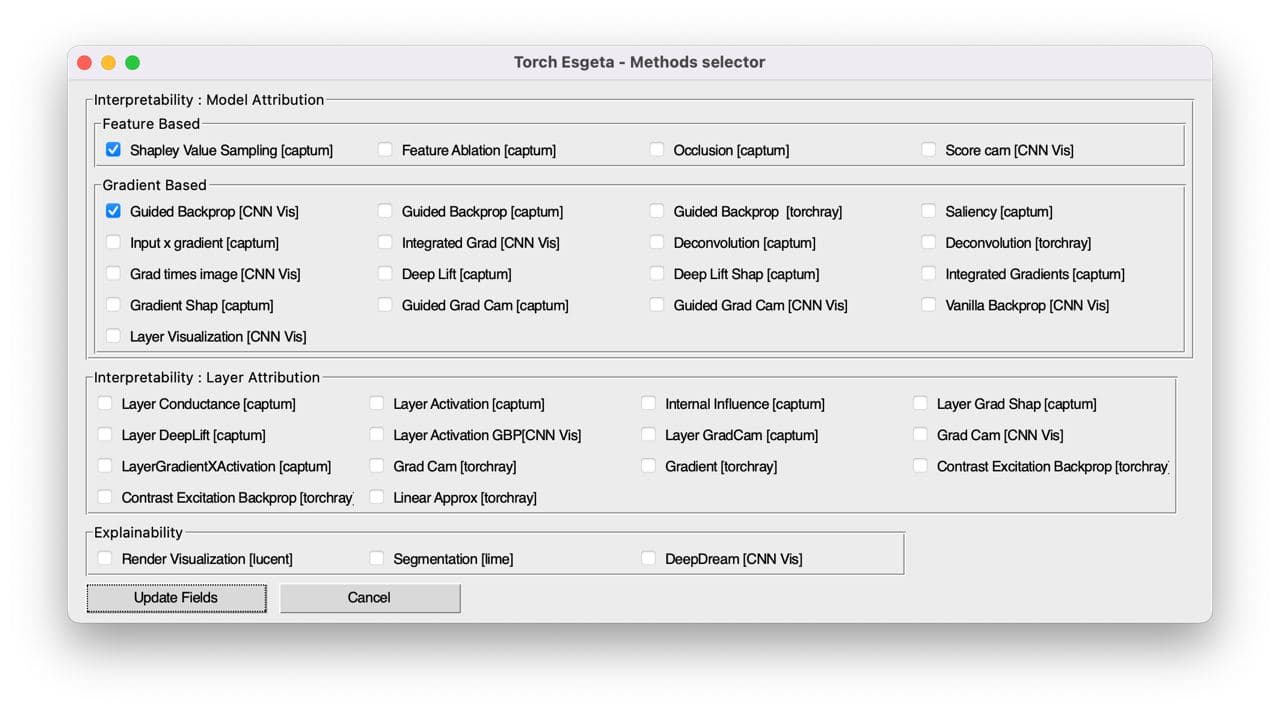}
			\captionsetup{justification=centering}
            \caption{GUI: Method selection. Choosing which methods are to be applied on given models.}
	        \label{fig:UI_2}
   \end{figure}

\begin{figure}
			\centering
			\includegraphics[width=0.6\textwidth]{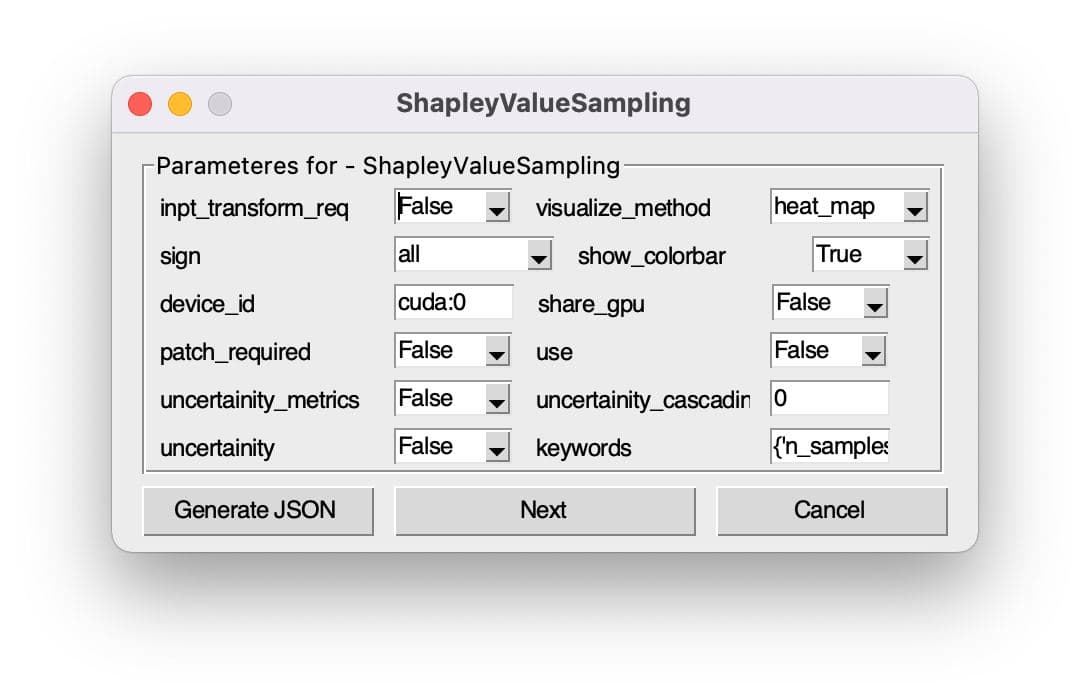}
			\captionsetup{justification=centering}
            \caption{GUI: Method-specific parameter selection for a chosen method.}
	        \label{fig:UI_3}
   \end{figure}

A Graphical User Interface (GUI) has been created to provide the end-user with an intuitive graphical layout to select the parameters and interpretability methods according to their requirement. The first screen shown in Figure \ref{fig:UI_1} provides the user with a dialogue box for parameter selection. The user can choose the model nature, model name, dataset and many more run-time parameters. Once these selections are made, the 'Select Methods' button will display the following dialogue box as shown in Figure \ref{fig:UI_2}. The users have a wide range of interpretability methods to choose from in this dialogue box. Once the interpretability methods are chosen, the method-specific parameter dialogue box is displayed to the user as in  Figure \ref{fig:UI_3}. The users can then choose the visualisation method, device id and many other parameters. On clicking the 'Next' button, the code will be executed in the back-end, and the output will be generated in the output path specified in Figure \ref{fig:UI_1}.

\subsection{Evaluation Methods}
\label{ch:evaluation}
In order to evaluate and compare the methods, both qualitative and quantitative aspects have been considered. 

\subsubsection{Qualitative evaluation}
For qualitative or visual evaluation, the cascading randomisation~\citep{adebayo2018sanity} technique was implemented, in which the model weights are randomised successively, from the top to the bottom layers. The learned weights of the layers are destroyed from top to bottom by this technique. While doing so, the interpretability and explainability techniques are applied to each state of randomisation. If the interpretability-explainability results are dependent upon the model's weights - how it is supposed to be - then the quality would decrease as the amount of randomisation increases. If they are not dependent, then the results will be unaffected by the randomisation - hence, the technique is not accurate or unsuitable for the current model. 

\subsubsection{Quantitative evaluation}
For the quantitative evaluation, the uncertainty method was chosen due to the unavailability of ground truth data for the attribution images. Two metrics have been used for computing the uncertainty values of the attribution methods:
Infidelity~\citep{yeh2019fidelity}] and Sensitivity~\citep{yeh2019fidelity}. These metrics can be used only on model attribution methods as of now. It is to be noted that the qualitative metrics can be used on top of cascading randomisation as an additional level of evaluation.

Infidelity\footnote{\label{foot} https://captum.ai/api/metrics.html} represents the expected mean-squared error between the explanation multiplied by a meaningful input perturbation and the differences between the predictor function at its input and perturbed input.
Sensitivity$^{\ref{foot}}$ measures the extent of explanation change when the input is slightly perturbed.

\section{Results}
\label{ch:Results and Analysis}
To evaluate the TorchEsegeta pipeline for segmentation models, a use-case of vessel segmentation was chosen. 

\subsection{Models} 
The segmentation network models chosen for this use case are from the DS6 paper~\citep{chatterjee2020ds6}.
The models are UNet, UNet MSS and UNet MSS with Deformation.
The difference between the UNetMSS and UNetMSS with Deformation is a change in the number of up-sampling and down-sampling layers and a modified activation function. UNetMSS performs downsampling five times using convolution striding and uses transported convolution for upsampling, and it includes instance normalisation and Leaky ReLU in the convolution blocks. On the contrary, the modified version applies four down-sampling layers using max-pool and performs upsampling using interpolation combined with Batch normalisation and ReLU in its convolution block.

For UNetMSS with Deformation, a small amount of variable elastic deformation is added at the time of training, along with each volume input.
 The authors have given a comprehensive overview of the method, and based on their results, one can see that UNetMSS with Deformation is the best performing model.

\subsection{Use Case Experiment}
\begin{figure}[!htb]
\centering
\includegraphics[width=1\textwidth]{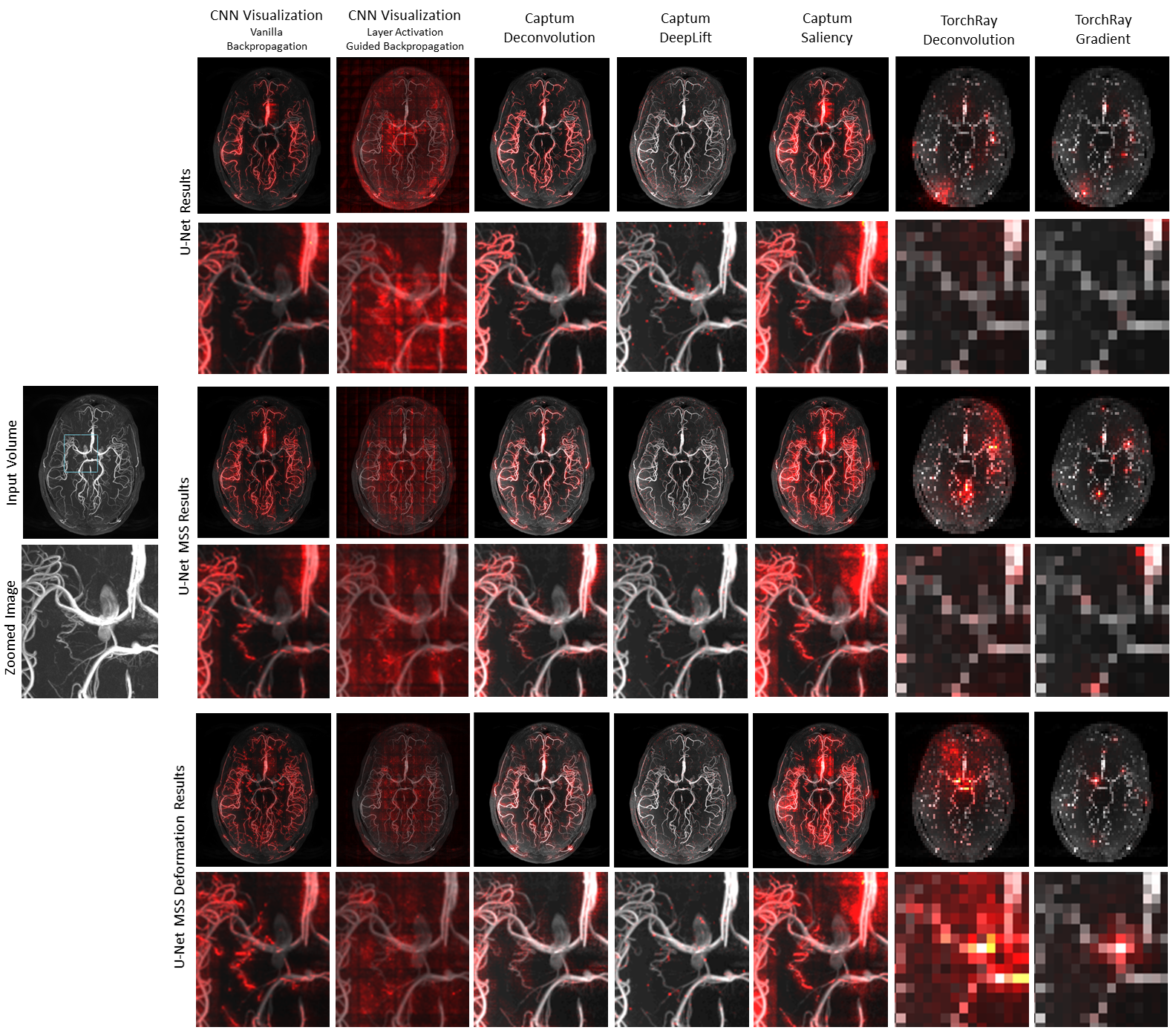}
\captionsetup{justification=centering}
\caption{Example of interpretability techniques compared across three models: U-Net, U-Net MSS and U-Net MSS with deformation, along with its corresponding zoomed image. Higher the intensity of red - higher the concentration of the focus of the network. Looking at the regions where the network did not focus, it can be understood which parts of the segmentation prediction might be wrong.}
\label{fig:compareTechModels}
\end{figure}

\begin{figure}[!htb] 
\centering
\includegraphics[width=0.8\textwidth]{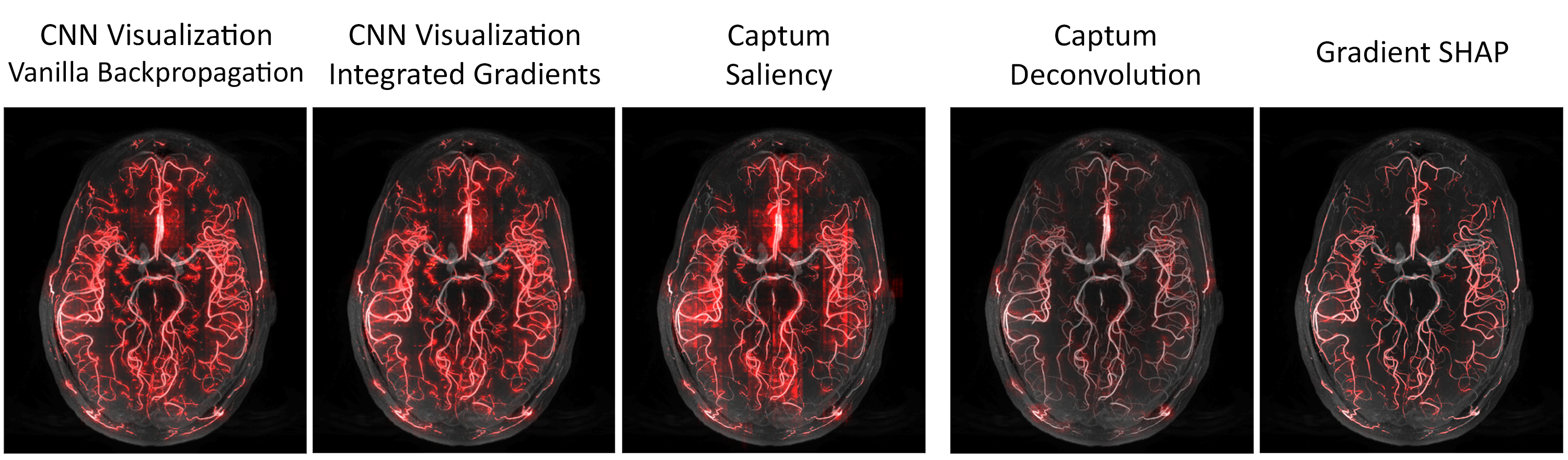}
\captionsetup{justification=centering}
\caption{Some of the similar interpretability results of U-Net MSS Deformation model. Higher the intensity of red - higher the concentration of the focus of the network. The results of CNN Visualization Vanilla Backpropagation, CNN Visualization Integrated Gradients, and Captum Saliency show similar focus areas: on anterior, posterior, right, and left regions of the brain. Captum Deconvolution and Gradient SHAP emphasise specifically on the cerebral artery itself, not the whole area.}
\label{fig:similarResultsUNetMssDef}
\end{figure}

As a use case study, the previously mentioned interpretability and explainability techniques were explored while analysing the results of a vessel segmentation model trained on Time-of-fight (TOF) Magnetic Resonance Angiogram (MRA) images of the human brain called DS6~\citep{chatterjee2020ds6}. The model automatically segments vessels from 3D 7 Tesla TOF-MRA images. A study about the attribution outputs of the different methods across the three models was conducted. The zoomed-in portions of the attribution images show even more closely the model's focus areas. Figure ~\ref{fig:compareTechModels} shows examples of interpretability techniques compared across the three models. Figure ~\ref{fig:similarResultsUNetMssDef} portrays similar interpretability results from U-Net MSS Deformation model. By observing the interpretability results, it can be understood where the network is focusing. In production, ground-truth segmentations are not available. By looking at the interpretability results, it might be possible to estimate in which regions the network might miss-predict: by looking at the areas where the network did not adequately focus. These interpretability results can assist radiologists to build trust in models which focus on correct regions. On the other hand, a model developer can benefit from looking at the wrong focus regions or regions with no focus - a knowledge that can be further used to improve the model architecture or the training process.

In all of these attribution maps, whichever pixel is highlighted by red represents high activation, indicating the most important pixels in the input according to the respective method, and the other region represents lower to no activation.

\subsection{Notable observations}
In Figure \ref{fig:compareTechModels}, the CNN Visualization Layer Activation Guided Backpropagation attribution map shows a lot of attributions, most of which are on parts other than the brain vessels, i.e. the region of interest. The respective zoomed images also confirm the observation. On the contrary, methods such as Captum Deeplift and TorchRay Gradient provide fewer attributions and miss out on major portions of the brain vessels. Torchray Deconvolution provides more attributions comparatively; however, most of the attributions are concentrated on certain areas of the brain. CNN Visualization Vanilla Backpropagation, Captum Deconvolution and Captum Saliency provide better results compared to the others - they provided more human-interpretable results. These methods mainly attribute on the region of interest. However, a look at the respective zoomed images would reveal that for Captum Saliency, there are attributions even in the areas surrounding the brain vessels. For the other two, the attributions are mainly on the brain vessels only. Figure \ref{fig:similarResultsUNetMssDef} shows a comparison of some of the better performing interpretability methods for model U-Net MSS Deformation model. 

\begin{figure}[!htb] 
\centering
\includegraphics[width=1.0\textwidth]{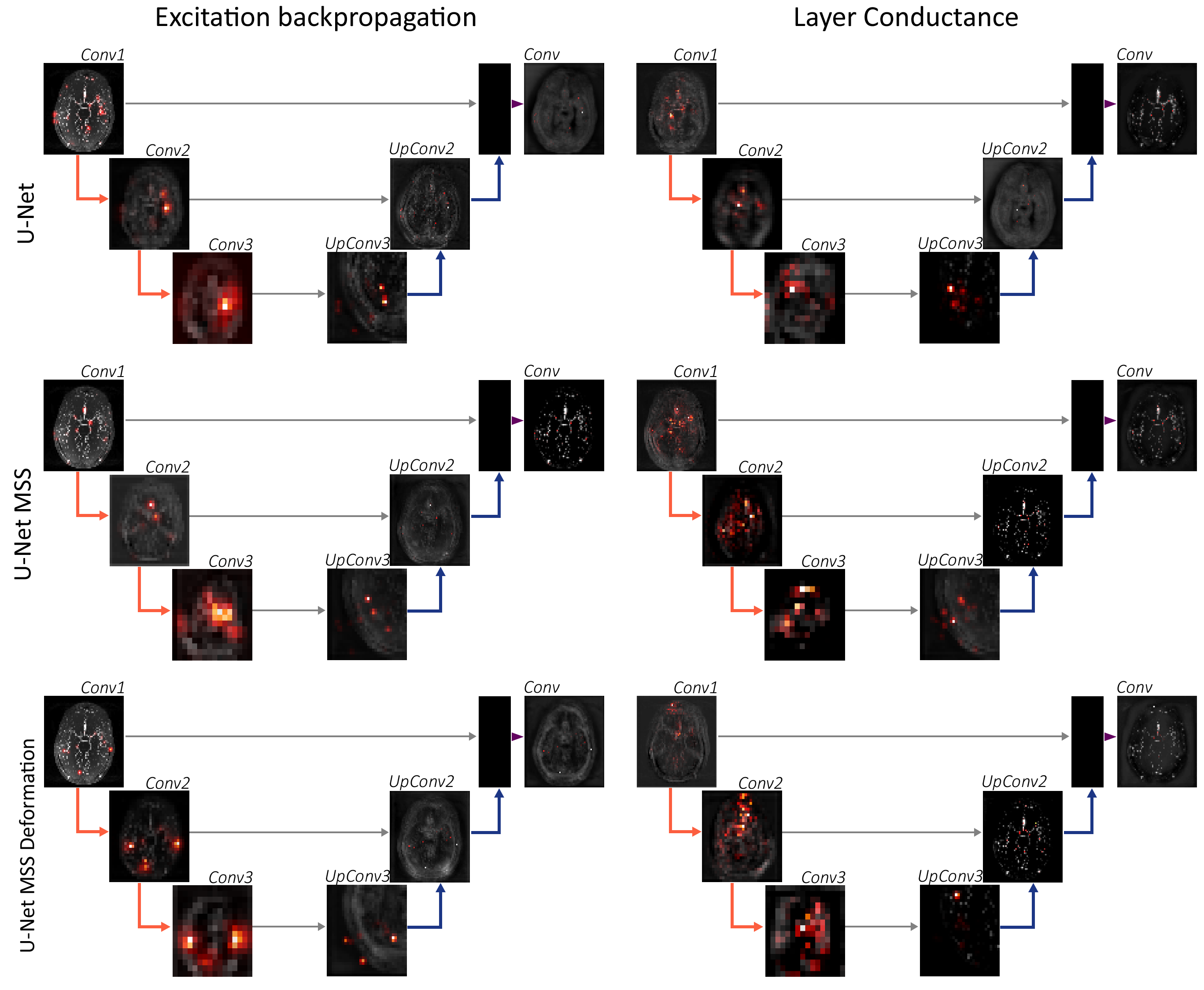}
\captionsetup{justification=centering}
\caption{Layer-based interpretability methods: Excitation Backpropagation and Layer Conductance, shown using representative UNets. Higher the intensity of red - higher the concentration of the focus of the network. This figure shows how the focus of the network changes in each layer for the three different models.}
\label{fig:layerResults_ExciteBackprop_LayConduct}
\end{figure}

\begin{figure}[!htb] 
\centering
\includegraphics[width=1.0\textwidth]{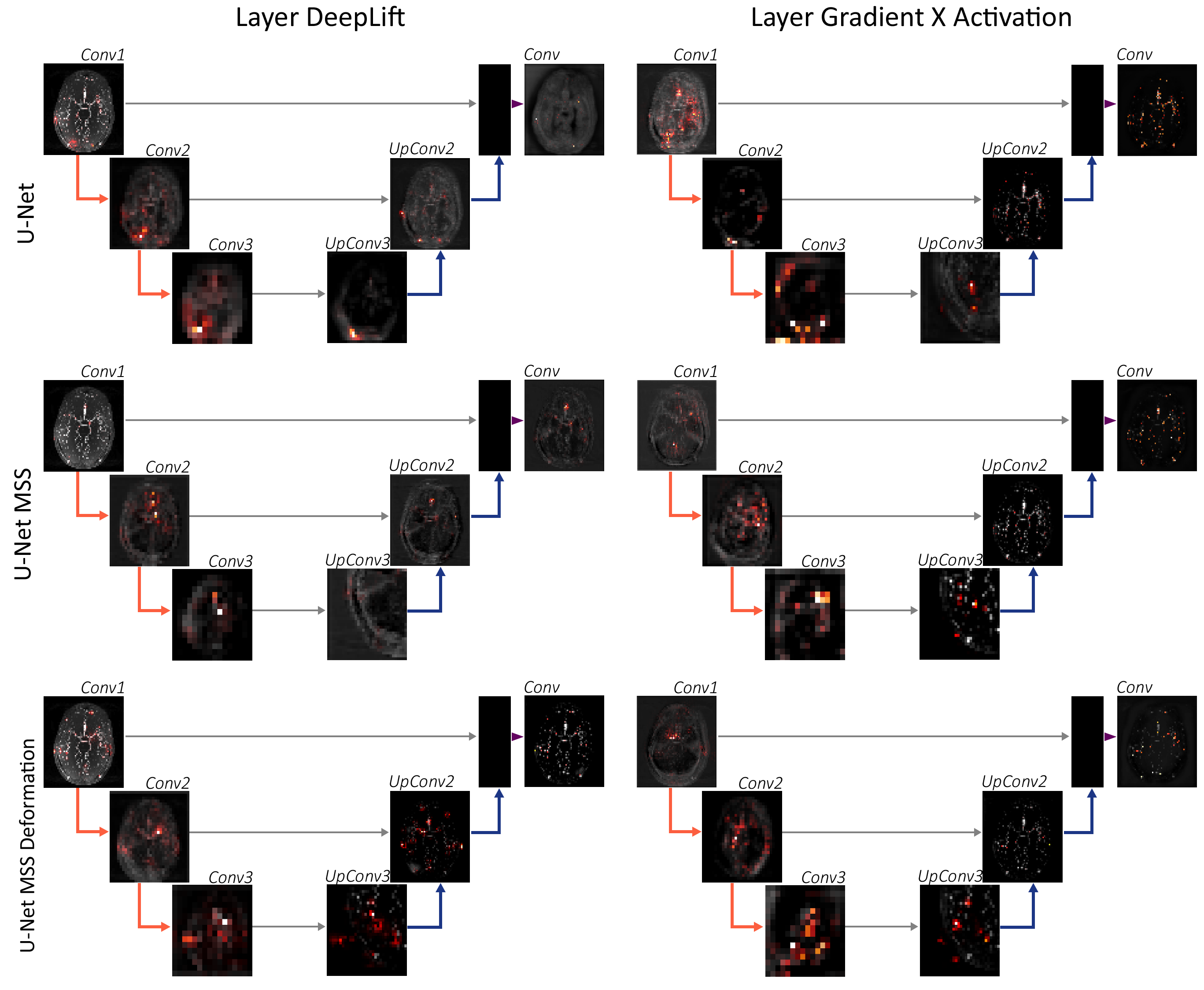}
\captionsetup{justification=centering}
\caption{Layer-based interpretability methods: Layer Deeplift and Layer Gradient X Activation, shown using representative UNets. Higher the intensity of red - higher the concentration of the focus of the network. This figure shows how the focus of the network changes in each layer for the three different models.}
\label{fig:layerResults_LayDeepLift_LayGradXAct}
\end{figure}

In Figure \ref{fig:layerResults_ExciteBackprop_LayConduct} and Figure \ref{fig:layerResults_LayDeepLift_LayGradXAct}, the layer-wise attributions for the three models are shown, for the methods: Excitation Backpropagation, Layer Conductance, Layer DeepLift and Layer Gradient X Activation. For all the methods, the maximum attributions are shown in the Conv3 layer. Among the three models, U-Net MSS Deformation seems to focus better than the other two, as it attributions can be seen all over the brain contrary to the other two (it is to be noted the vessels are present all over the brain, and not focused in a specific region). The network focus shifts while going from the first to the last layer of the models. In the initial layers, the focus is more on selective regions of the brain. However, towards the deeper layers like Conv3, the focus is almost on the entire brain.

\subsection{Evaluation}
\begin{figure}[!htb]  
\centering
\includegraphics[width=0.6\textwidth]{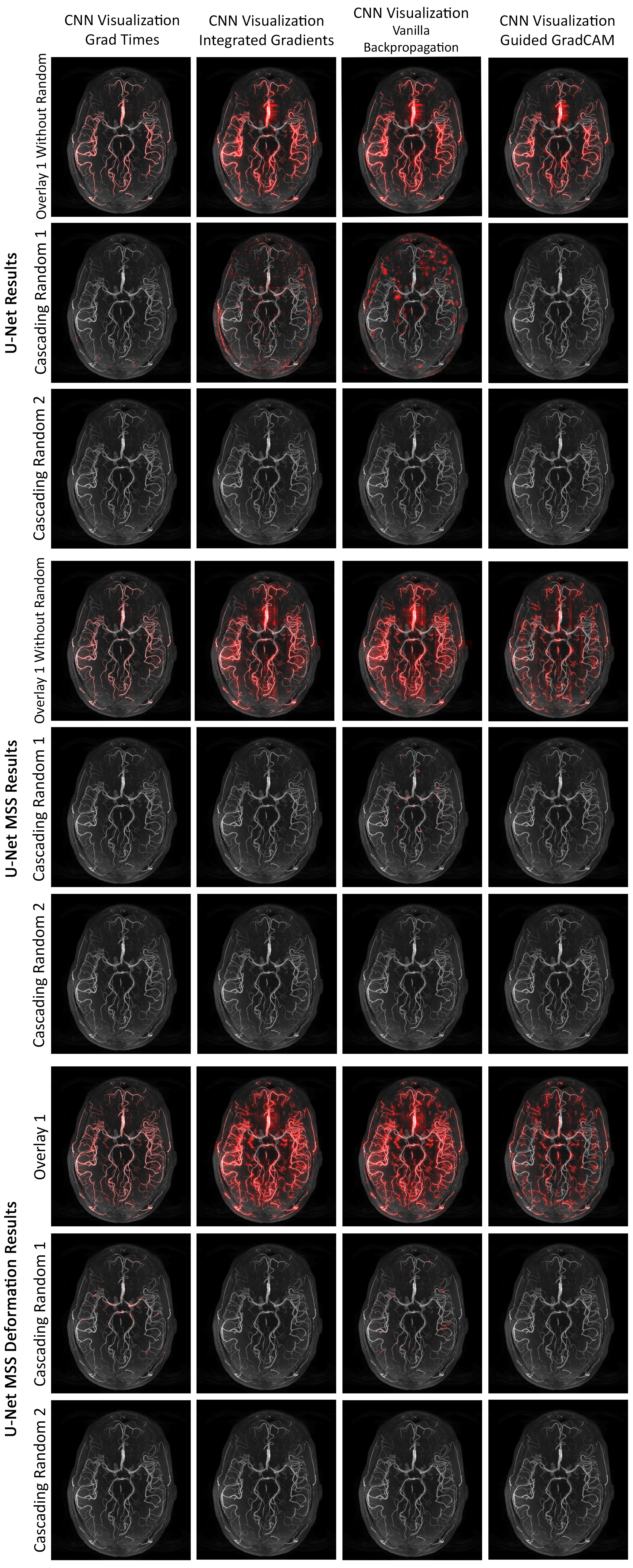}
\captionsetup{justification=centering}
\caption{Example of cascading randomisation outputs of the three models: U-Net, U-Net MSS and U-Net MSS with deformation. The outputs of all models and methods show the focus without randomisation fading through cascading randomisation 1 and 2. This implies that the output of these interpretability methods relay upon the weights of the network and not just random predictions. CNN Visualization Guided GradCAM shows the strongest dependency on the weights as even with cascading 1, all the attributions disappeared. Comparing the attributions for UNet against the other two models, it can be observed that the dependency on weights is stronger for the other models.}
\label{fig:cascadeRandom}
\end{figure}

A comparative study of the cascading randomisation technique is shown for the outputs of 4 interpretability methods for all the models in Figure ~\ref{fig:cascadeRandom}.

Moreover, to show the functionality of the quantitative evaluation part of the pipeline, a few methods were compared quantitatively using infidelity and sensitivity, the scores are shown in Table~\ref{tab:infidelity}~and~\ref{tab:sensitivity }.

\begin{table}[h!]
\captionsetup{justification=centering}
\caption{Infidelity Scores}
\centering
\begin{tabular}{c|c|c|c}
                & UNet     & UNetMSS  & UNetMSS\_Deform \\ \hline
Guided Backprop & \num{5.87e-17} & \num{2.08e-17} & \num{2.59e-18}        \\
Deconvolution   & \num{3.20e-16} & \num{2.62e-16} & \num{1.48e-16}        \\
Saliency        & \num{1.95e-15} & \num{1.54e-15} & \num{1.23e-16}        \\ \hline
\end{tabular}
\label{tab:infidelity}
\end{table}

\begin{table}[h!]
\captionsetup{justification=centering}
\caption{Sensitivity Scores}
\label{tab:sensitivity }
\centering
\begin{tabular}{c|c|c|c}
                & UNet     & UNetMSS  & UNetMSS\_Deform \\ \hline
Guided Backprop & \num{1.156} & \num{0.917} & \num{0.831}        \\
Deconvolution   & \num{1.210} & \num{1.188} & \num{1.140}        \\
Saliency        & \num{1.171} & \num{1.197} & \num{1.153}        \\ \hline
\end{tabular}
\end{table}
\section{Discussion}
\label{ch:conclusions}
In this work, various interpretability and explainability methods were adopted for segmentation models and were used to interpret the network. A pipeline has been developed and made public on GitHub\footnoteref{git}, TorchEsegeta, comprised of those interpretability and explainability methods that can be applied on 2D or 3D image-based deep learning models for classification and segmentation. The pipeline was experimented with using three models – UNet, UNetMSS and the UNetMSS-Deform, for the task of vessel segmentation from MRAs, using interpretability methods. The evaluation of the methods have been done qualitatively using cascading randomisation and quantitatively using evaluation metrics. It is worth mentioning that several explainability methods are part of the pipeline, but they were not evaluated with any use-case scenario during this research. 

This pipeline can be used by data scientists to improve their models by tweaking their models based on the shown interpretability and explainability - to improve the reasoning of the models, in turn improving the performance. Moreover, they can use this pipeline to show it to the model's users, for them to have trust in the model while using them in high-risk situations. On the other hand, this pipeline can be used by expert decision-makers, like clinicians, as a decision support system - by understanding the reasoning done by the models, they can get assistance in decision making.

It is noteworthy that the current pipeline can be extended for reconstruction purposes, and new interpretability and explainability methods can be added to the existing pipeline. Ground truth based pixel-wise interpretability for segmentation models can be implemented, which will add a new dimension to the existing work. Nevertheless, the real evaluation of the interpretations and explanations can only be done by the domain experts - in this case of vessel segmentation, by the clinicians - who can judge whether these results are actually useful or not. This step of the evaluation was not performed under the scope of the current work and will be performed in the near future.

\section*{Author Contributions}
Conceptualisation, S.C.; Literature Survey, A.S. and R.N.R.; Architecture Design, S.C. and A.D.; Pipeline Development, A.D., C.M. and B.M.; Experiments, A.D.; Quantitative Evaluation, M.V.; GUI Development, B.M.; Visualisation, C.S.; Writing - original draft, S.C., C.M. and R.N.R.; Writing - review and editing, S.C., O.S. and A.N. All authors have read and agreed to the submitted version of the manuscript.

\section*{Funding}

This work was in part conducted within the context of the International Graduate School MEMoRIAL at Otto von Guericke
University (OVGU) Magdeburg, Germany, kindly supported by the European Structural and Investment Funds (ESF) under the
programme "Sachsen-Anhalt WISSENSCHAFT Internationalisierung" (project no. ZS/2016/08/80646).

\section*{Conflicts of Interest}
The authors declare no conflict of interest.


\begin{adjustwidth}{-\extralength}{0cm}

\reftitle{References}
\bibliography{mybibfile}

%


\end{adjustwidth}
\end{document}